\documentclass[10pt,twocolumn,letterpaper]{article}

\usepackage{cvpr}
\usepackage{times}
\usepackage{epsfig}
\usepackage{graphicx}
\usepackage{amsmath}
\usepackage{amssymb}
\usepackage{bbm}
\usepackage{subcaption}
\usepackage{multirow}
\usepackage{adjustbox}
\usepackage[dvipsnames]{xcolor}

\usepackage{algorithm}
\usepackage{algorithmic}

\usepackage[pagebackref=true,breaklinks=true,letterpaper=true,colorlinks,bookmarks=false]{hyperref}

\DeclareMathOperator{\E}{\mathbb{E}}

\newcommand*{\ShowNotes}{}

\definecolor{darkred}{rgb}{0.7,0.1,0.1}
\definecolor{darkgreen}{rgb}{0.1,0.7,0.1}
\definecolor{cyan}{rgb}{0.7,0.0,0.7}
\definecolor{dblue}{rgb}{0.2,0.2,0.8}
\definecolor{maroon}{rgb}{0.76,.13,.28}
\definecolor{burntorange}{rgb}{0.81,.33,0}

\ifdefined\ShowNotes
  \newcommand{\colornote}[3]{{\color{#1}\bf{#2: #3}\normalfont}}
\else
  \newcommand{\colornote}[3]{}
\fi

\cvprfinalcopy 

\def\cvprPaperID{5192} 

\ifcvprfinal\fi
\begin{document}

\teaser{
  \centering
    \begin{tabular}{ccccccccccc}

   \includegraphics[height=0.09\textwidth]{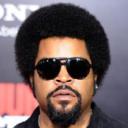}&
    \includegraphics[height=0.09\textwidth]{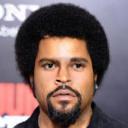}& 
    \includegraphics[height=0.09\textwidth]{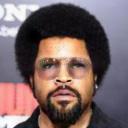}&

     \includegraphics[height=0.09\textwidth]{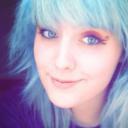}& 
    \includegraphics[height=0.09\textwidth]{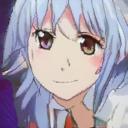}&
    \includegraphics[height=0.09\textwidth]{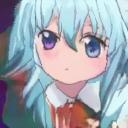}& 
    \includegraphics[height=0.09\textwidth]{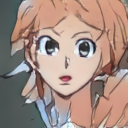}&
    
    \includegraphics[height=0.09\textwidth]{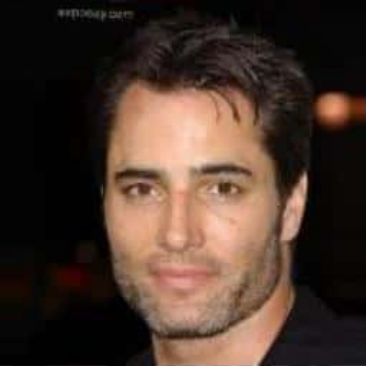}&
    \includegraphics[height=0.09\textwidth]{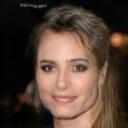}&
    \includegraphics[height=0.09\textwidth]{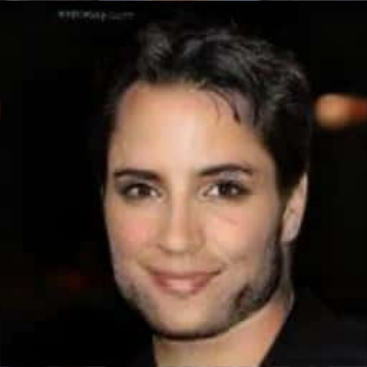}&
\\

    \includegraphics[height=0.09\textwidth]{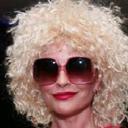}& 
    \includegraphics[height=0.09\textwidth]{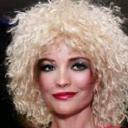}&
    \includegraphics[height=0.09\textwidth]{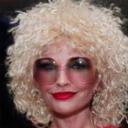}&
    
    \includegraphics[height=0.09\textwidth]{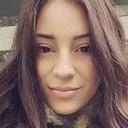}& 
    \includegraphics[height=0.09\textwidth]{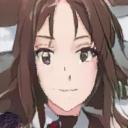}&
    \includegraphics[height=0.09\textwidth]{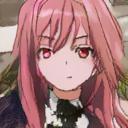}& 
    \includegraphics[height=0.09\textwidth]{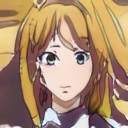}&

    \includegraphics[height=0.09\textwidth]{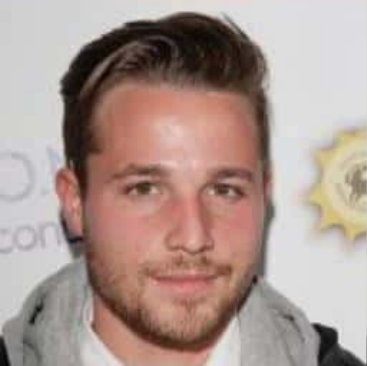}&
    \includegraphics[height=0.09\textwidth]{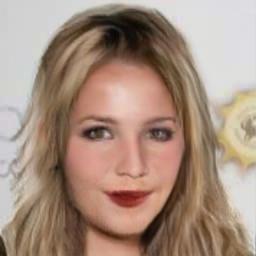}& 
    \includegraphics[height=0.09\textwidth]{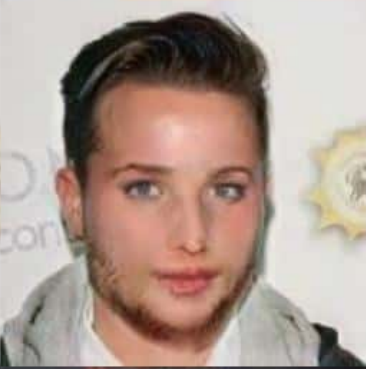}&
\\

    \includegraphics[height=0.09\textwidth]{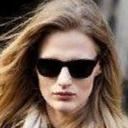}& 
    \includegraphics[height=0.09\textwidth]{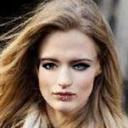}&
    \includegraphics[height=0.09\textwidth]{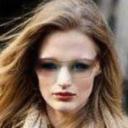}&
    
    \includegraphics[height=0.09\textwidth]{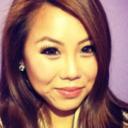}& 
    \includegraphics[height=0.09\textwidth]{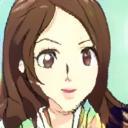}&
    \includegraphics[height=0.09\textwidth]{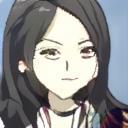}& 
    \includegraphics[height=0.09\textwidth]{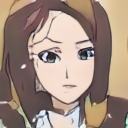}&
    
    \includegraphics[height=0.09\textwidth]{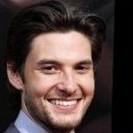}&
    \includegraphics[height=0.09\textwidth]{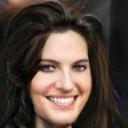}&
    \includegraphics[height=0.09\textwidth]{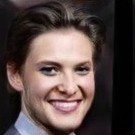}&

    \\
    input & ours & \cite{siddiquee2019learning} & input & ours-1 & ours-2 & \cite{DBLP:journals/corr/abs-1907-10830} & input & ours & \cite{StarGAN} \\
    \multicolumn{3}{c}{(a) glasses removal} &
    \multicolumn{4}{c}{(b) selfie to anime} &
    \multicolumn{3}{c}{(c) male to female}  \\
    \end{tabular}
    \caption{
    {\bf Translation results for three applications.}
    Our method is unidirectional (cycles are unnecessary) and multi-modal (multiple results are generated for a given input, e.g.~(b)).
    The results are compared to SOTA results and shown to outperform them.
    In~(a) our method completely removes the glasses; in~(b) the shape of the face is well maintained, and in~(c) the women look more "feminine", e.g., no beard leftovers.
    More results \& comparisons can be found later in the paper.
   }
  \label{fig:teaser}
}


\title{Breaking the cycle---Colleagues are all you need}
\author{Ori Nizan\\
Technion, Israel\\
{\tt\small snizori@campus.technion.ac.il}
\and
Ayellet Tal\\
Technion, Israel\\
{\tt\small ayellet@ee.technion.ac.il}
}

\maketitle

\begin{abstract}
This paper proposes a novel approach to performing image-to-image translation between unpaired domains.
Rather than relying on a cycle constraint, our method takes advantage of collaboration between various GANs.
This results in a multi-modal method, in which multiple optional and diverse images are produced for a given image.
Our model addresses some of the shortcomings of classical GANs:
(1) It is able to remove large objects, such as glasses.
(2) Since it does not need to  support the cycle constraint, no irrelevant traces of the input are left on the generated image.
(3) It manages to  translate between domains that require large shape modifications.
Our results are shown to outperform those generated by state-of-the-art methods for several challenging applications.
Code Available at https://github.com/Onr/Council-GAN
\end{abstract}

\section{Introduction}
\label{sec:Introduction}

Mapping between different domains is inline with the human ability to find similarities between features in distinctive, yet associated, classes.
Therefore it is not surprising that image-to-image translation has gained a lot of attention in recent years.
Many applications have been demonstrated to benefit from it, yielding  beautiful results.

In unsupervised settings, where no paired data is available, shared latent space and cycle-consistency assumptions have been utilized~\cite{Anoosheh2017ComboGANUS, StarGAN, Chu2017,Hua2017UnsupervisedCI,  Huang2018MultimodalTranslation, kim2017,Liu2017, royer2017xgan,YiDualGAN, Zhu2017UnpairedNetworks}. 
Despite the successes \& benefits, previous methods might suffer from some drawbacks.
In particular, oftentimes, the cycle constraint might cause the preservation of source domain features, as can be seen for example, in Figure~\ref{fig:teaser}(c), where facial hair remains on the faces of the women. 
This is due to the need to go back and forth through the cycle.
Second, as discussed in~\cite{DBLP:journals/corr/abs-1907-10830}, sometimes the methods are unsuccessful for image translation tasks with large shape change, such as in the case of the anime in Figure~\ref{fig:teaser}(b).
Finally, as explained in~\cite{siddiquee2019learning}, 
it is still a challenge to completely remove large objects, like glasses, from the images, and therefore this task is left for their future work (Figure~\ref{fig:teaser}(a)).

We propose a novel approach, termed {\em Council-GAN}, which handles these challenges.
The key idea is to rely on "collegiality" between GANs, rather than utilizing a cycle.
Specifically, instead of using a single pair of a generator/discriminator "experts", it utilizes the collective opinion of a group of pairs (the {\em council}) and leverages the variation between the results of the generators.
This leads to a more stable and diverse domain transfer.

To realize this idea, we propose to train a council of multiple council members, requiring them to learn from each other.
Each generator in the council gets the same input from the source domain and will produce its own output.
However, the outputs produced by the various generators should have some common denominator.
For this to happen across all images, the generators have to find common features in the input, which are used to generate their outputs. 
Each discriminator learns to distinguish between the generated images of its own generator and images produced by the other generators.
This forces each generator to converge to a result that is agreeable by the others.
Intuitively, this convergence assists to  maximize the mutual information between the source domain and the target domain, which explains why the generated images maintain the important features of the source images.

We demonstrate the benefits of our approach for several applications, including glasses removal, face to anime translation, and male to female translation.
In all cases we achieve state-of-the-art results.

Hence, this paper makes the following contributions:
\begin{enumerate}
    \item 
    We introduce a novel model for unsupervised image-to-image translation, whose key idea is collaboration between multiple generators.
    Conversely to most recent methods, our model  avoids cycle-consistency constraints altogether. 
    \item
    Our council manages to achieve state-of-the-art results in a variety of challenging applications.
\end{enumerate}

\section{Related work}
\label{sec:related work}
\noindent
{\bf Generative adversarial networks (GANs). }
Since the introduction of the GAN framework~\cite{Goodfellow2014}, it has been demonstrated to achieve eye-pleasing results in numerous applications.
In this framework, a generator is trained to fool a discriminator, whereas the latter attempts  to distinguish between the generated samples and real samples. 
A variety of modifications have been proposed in recent years in an attempt to improve  GAN's results; see~\cite{Arjovsky2017WassersteinNetworks,Denton2015DeepNetworks,dosovitskiy2016generating,Huang2016StackedNetworks,Karras2018ProgressiveVariation,Mao2017LeastNetworks,RadfordDCGANUNSUPERVISEDNETWORKS,   Rosca2017VariationalNetworks,Salimans2016ImprovedGANs,Tolstikhin2018WassersteinAuto-Encoders,ZhangStackGAN:Networks} for a few of them.

We are not the first to propose the use of multiple GANs~\cite{Durugkar2016GenerativeNetworks,Ghosh2018Multi-agentNetworks, QuanHoangTuDinhNguyenTrungLe2018, Juefei-Xu2017GangRanking}.
However, previous approaches differ from ours both in their architectures and in their goals.
For instance, some of previous architectures consist  of multiple discriminators and a single generator; conversely, some propose to have a key discriminator that can evaluate the generators' results and improve them.
We propose a novel architecture to realize the concept of a council, as described in Section~\ref{sec:council}.
Furthermore, the goal of other approaches is either to push each other apart, to create diverse solutions, or to improve the results.
Our council attempts to find the commonalities between the the source and target domains.
By requiring the council members to "agree" on each other's results, they in fact learn to focus on the common traits of the domains.

\noindent
{\bf Image-to-image translation.} 
The aim is to learn a mapping from a source domain to a target domain. 
Early approaches adopt a supervised framework, in which the model learns paired examples, for instance using a conditional GAN to model the mapping function~\cite{ Isola2016Image-to-ImageNetworks, wang2018pix2pixHD, Zhu2017}. 

Recently, numerous methods have been proposed, which use unpaired examples for the learning task and produce highly impressive results; see for example ~\cite{Berthelot2017BEGAN:Networks,Gatys2016,Huang2018MultimodalTranslation,kim2017,Lee_2018_ECCV,Liu2017,siddiquee2019learning,Zhu2017UnpairedNetworks}, out of a truly extensive literature.
This approach is vital  to applications for which paired data is unavailable or difficult to gain.
Our model belongs to the class of GAN models that do not require paired training data.

A major concern in the unsupervised approach is the type of properties of the source domain that should be preserved.
Examples include
pixel values~\cite{Shrivastava2017LearningTraining}, 
pixel gradients~\cite{bousmalis2017unsupervised},  
pairwise sample distances~\cite{benaim2017one}, and recently mostly
cycle consistency~\cite{kim2017,YiDualGAN,Zhu2017UnpairedNetworks}.
The latter enforces the constraint that translating an image to the target domain and back, should obtain the original image.
Our method avoids using cycles altogether.
This has the benefit of bypassing unnecessary constraints on the generated output, and thus avoiding to preserve hidden information~\cite{Chu2017}.

Most existing methods lack diversity in the results. 
To address this problem, some methods propose to produce multiple outputs for the same given image~\cite{Huang2018MultimodalTranslation,Lee_2018_ECCV}.
Our method enables image translation with diverse outputs, however it does so in a manner in which all GANs in the council "acknowledge" to some degree each other's output.

\noindent
{\bf Ensemble methods.}
These methods use multiple learning algorithms, trained individually\cite{opitz1999popular, polikar2006ensemble, rokach2010ensemble}, whose predictions are combined.
They seek to  promote diversity among the models they combine.
Conversely, we require the council to learn together and "converge" to agreeable solutions.

\section{Model}
\label{sec:council}

This section describes our proposed model, which addresses the drawbacks described in Section~\ref{sec:Introduction}.
Our model consists of a set, termed a {\em council}, whose members influence each other's results.
Each member of the council has one generator and a couple of discriminators, as described below.
The generators need not converge to a specific output; instead, each produces its own results, jointly generating a diverse set of results.
During training, they take into account the images produced by the other generators.
Intuitively, the mutual influence enforces the generators to focus on joint traits of the images in the source domain, which could be matched to those in the target domain.
For instance, in Figure~\ref{fig:teaser}, to transform a male into a female, the generators focus on the structure of the face, on which they can all agree upon.
Therefore, this feature will be preserved, which can explain the good results.

Furthermore, our model avoids cycle constraints.
This means that there is no need to go in both directions between the source domain and the target domains.
As a result, there is no need to leave traces on the generated image (e.g. glasses) or to limit the amount of change (e.g. anime). 

\begin{figure}[tb]
\centering
\includegraphics[width=0.4\textwidth]{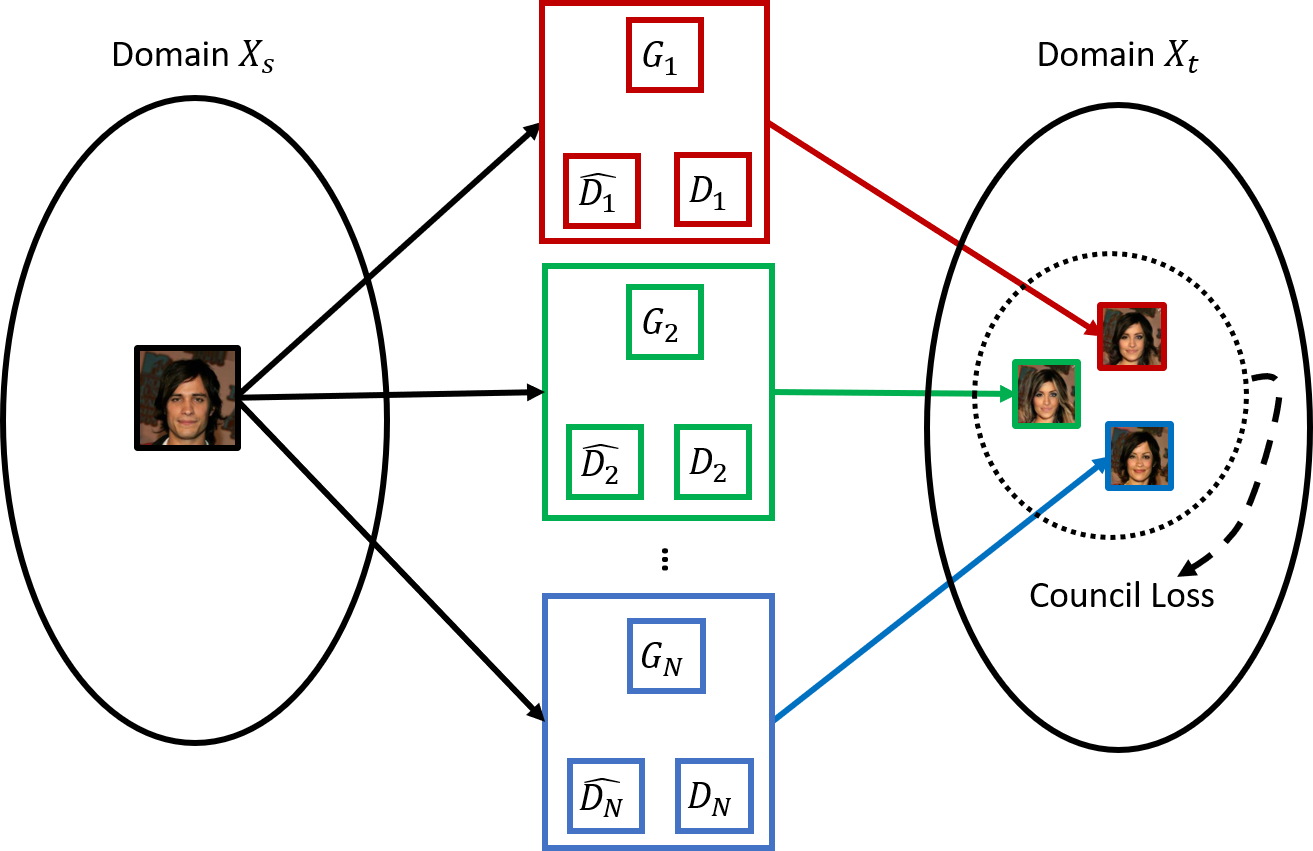}
\caption{{\bf General approach.} 
The council consists of triplets, each of which contains a generator and two discriminators: 
~\(D_i\) distinguishes  between the generator's output and real examples, whereas~\(\hat{D_i}\) distinguishes between images produced by $G_i$ and images produced by other generators in the council.
\(\hat{D_i}\) is the reason that the each of the generators converges to a result that is agreed-upon by all other members of the council. 
}
\label{fig:council}
\end{figure}

To realize this idea, we define a council of $N$ members as follows (Figure~\ref{fig:council}).
Each member $i$ of the council is a triplet, whose components are a single generator $G_i$ and two discriminators \(D_i\) \& \(\hat{D_i}\), $1 \leq i \leq N$.
The task of discriminator~\(D_i\) is to distinguish between the generator's output and real examples from the target domain, as done in any classical GAN.
The goal of discriminator~\(\hat{D_i}\) is to distinguish between images produced by \(G_i\) and images produced by the other generators in the council. 
This discriminator is the core of the model and this is what differentiates our model from the classical GAN model.
It enforces the generator to converge to images that could be acknowledged by all council members---images that share similar features.

The loss function of $D_i$ is the classical adversarial loss of~\cite{Mao2017LeastNetworks}.
Hereafter, we focus on the loss function of  \(\hat{D_i}\), which makes the outputs of the various generators share common traits, while still maintain diversity.
At every iteration, \(\hat{D_i}\) gets as input pairs of (input,output) from all the generators in the council.
Rather than distinguishing between real \& fake, $\hat{D_i}$'s distinguishes between the result of "my-generator" and the result of "another-generator".
Hence, during training, $G_i$ attempts to minimize the distance between the outputs of the generators.
Note that getting the input and not only the output is important to make the connection, for each pair, between the features of the source image and those of the generated image. 

Let $X_s$ be the source domain and $X_t$ be the target domain.
In our model we have $N$ mappings $G_i: X_s \rightarrow X_t$.
Given an image $x \in X_s$, a straightforward adaptation of the classical adversarial loss to our case would be: 
 \begin{eqnarray}
 \label{eq:loss_naive}
     Naive\_council\_loss_i(G_i, \hat{D_i}, \{G_j\}_{j\neq i}, X_s)= &&\\
       \E_{x \sim p(X_s)}\sum_{j\neq i}[log(1-\hat{D_i}(G_i(x),x))&& \nonumber\\
     +log(\hat{D_i}(G_j(x),x))],&& \nonumber
 \end{eqnarray}
where $G_i$ tries to generate images $G_i(x)$ that look similar to images from domains $G_j(x)$ for $j \neq i$. 
In analogy to the classical adversarial loss, in Equation~\eqref{eq:loss_naive}, both terms should be minimized, where the left term learns to "identify" its corresponding generator $G_i$ as "fake" and the right term learns to "identify" the other generators as "real".

\begin{figure}[tb]
\centering
\includegraphics[width=0.42\textwidth]{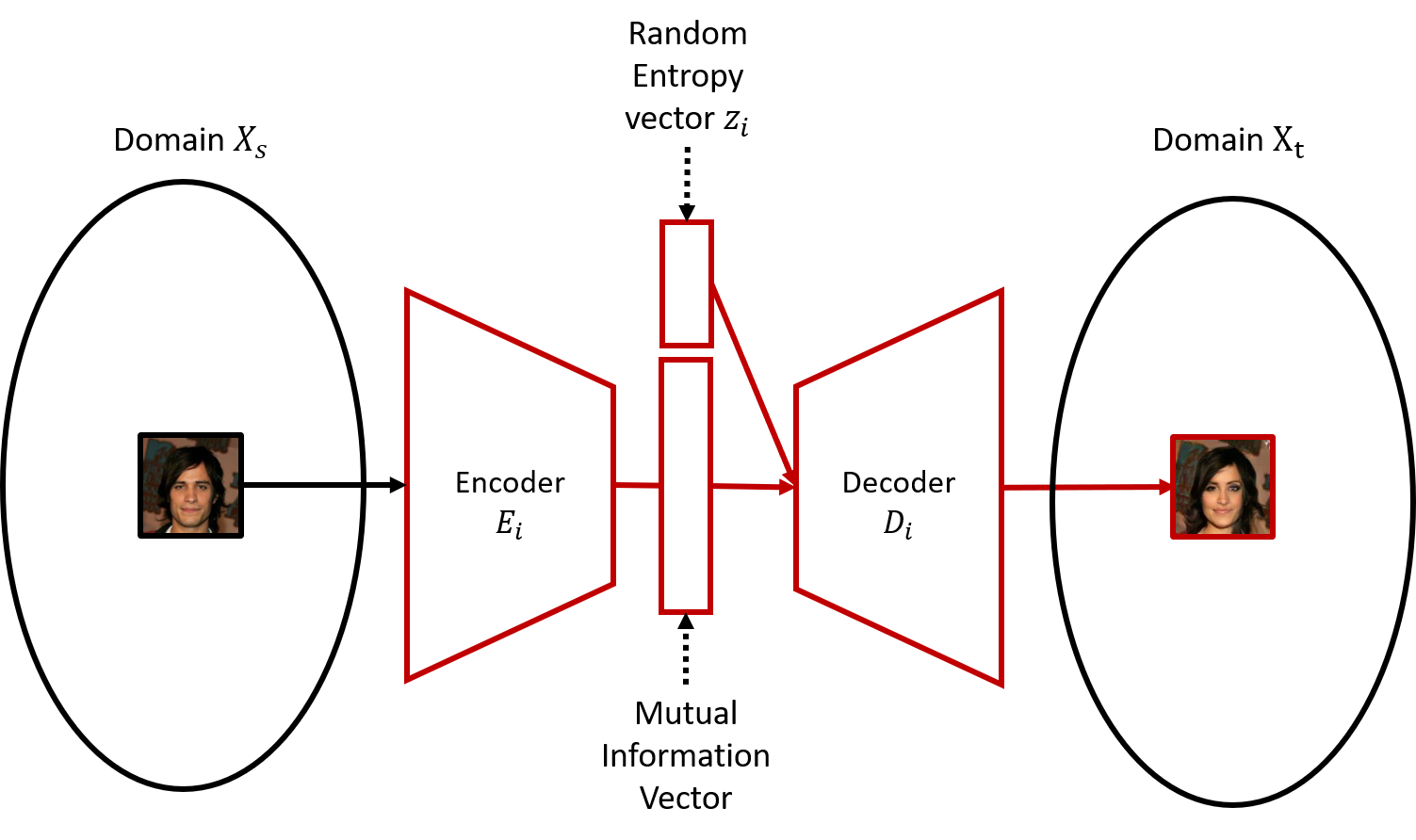}
\caption{{\bf Zoom into the generator $G_i$.}
Our generator is an auto-encoder architecture, which is similar to that of~\cite{Huang2018MultimodalTranslation}.
The encoder consists of several strided convolutional layers followed by residual blocks.
The decoder gets the encoded image (termed the {\em mutual information vector}), as well as a random entropy vector.
The latter  may be interpreted as encoding the leftover information of the target domain.
The decoder uses a MLP to produce a set of AdaIN parameters for the random entropy vector~\cite{Huang2017ArbitraryST}.
}
\label{fig:council_zoom}
\end{figure}

\begin{figure*}[tb]
\centering
\begin{tabular}{cc}
\includegraphics[width=0.5\textwidth]{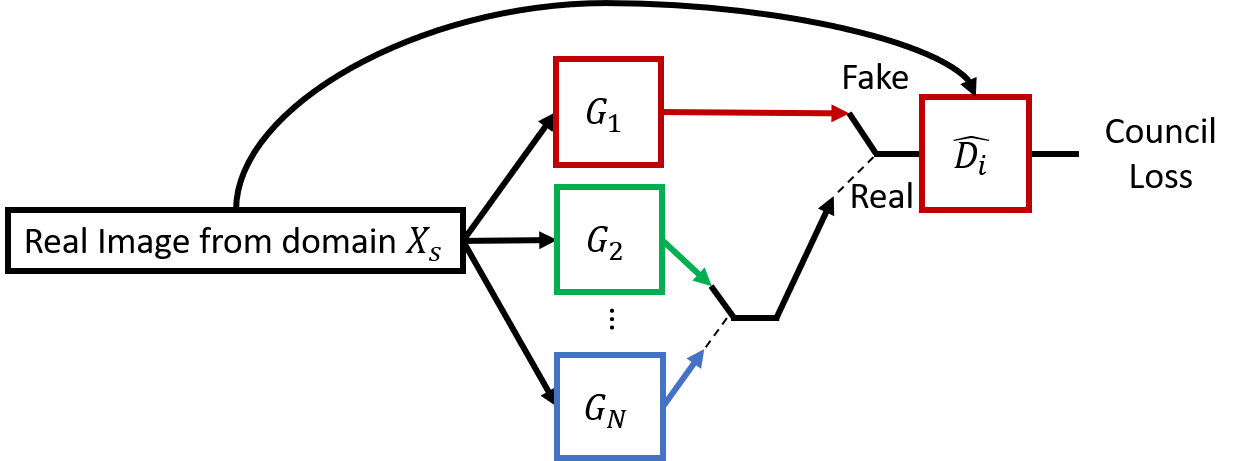}&
\includegraphics[width=0.45\textwidth]{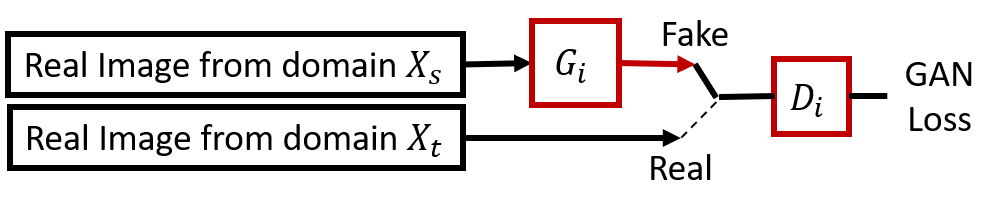}\\
(a) Council discriminator $\hat{D}_i$ &
(b) GAN discriminator $D_i$ 
\end{tabular}
\caption{
{\bf Differences \& similarities between  $\hat{D}_i$  and  $D_i$.}
While the GAN discriminator distinguishes between "real" and "fake" images, the council discriminator distinguishes between outputs of its own generator and those produced by other generators.
Furthermore, while the GAN's discriminator gets as input only the generator's output, the council's discriminator gets also the generator's input. 
This is because  we wish the generator to produce a result that bares similarity to the input image, and not only one that looks real in the target domain. 
}
\label{fig:discriminators}
\end{figure*}

To allow multimodal translation, we encode the input image, as illustrated in Figure~\ref{fig:council_zoom}, which zooms into the structure of the generator~\cite{Huang2018MultimodalTranslation}.
The encoded image should carry useful (mutual) information between domains $X_s$ and $X_t$. 
Let $E_i$ be the $i^{th}$ encoder for the source image and let $z_i$ be the  $i^{th}$ random entropy vector, associated with the  $i^{th}$ member of the council,  $1 \leq i \leq N$.
$z_i$ enables each generator to generate multiple diverse results.
Equation~\eqref{eq:loss_naive} is modified so as to get an encoded image (instead of the original input image) and the random entropy vector.
The loss function of $\hat{D_i}$ is then defined as:
\begin{eqnarray}
\label{eq:loss_council}
Council\_loss_i(G_i, \hat{D_i}, \{G_j\}_{j\neq i}, X_s, z_i, \{E_j\}_{1 \leq j \leq N})= && \\
\E_{x \sim p(X_s)}
\sum_{j\neq i}[log(1-\hat{D_i}(G_i(E_i(x),z_i),x)) && \nonumber \\
+ log(\hat{D_i}(G_j(E_j(x),\alpha z_j),x))]. &&\nonumber
\end{eqnarray}
Here, the loss function gets, as additional inputs, all the encoders and vector $z_i$.
$\alpha$ controls the size of the sub-domain of the other generators, which is important in order to converge to "acceptable" images.

Figure~\ref{fig:discriminators} illustrates the differences and the similarities between discriminators $D_i$ and $\hat{D}_i$.
Both should distinguish between the generator's results and other images; in the case of $D_i$ the other images are real images from the target domain, whereas in the case of $\hat{D}_i$, they are images generated by other generators in the council.
Another fundamental difference is their input: $\hat{D}_i$ gets not only the generator's output, but also its input. 
This aims at producing a resulting image that has common features with the input image.

\noindent
{\bf Final loss.} 
For each member of the council, we jointly train the generator (assuming the encoder is included) and the discriminators to optimize the final objective.
In essence, $G_i$, $D_i$, \& $\hat{D}_{i}$ play a three-way min-max-max game with a value function $V(G_i,  D_i,\hat{D}_{i})$:

\begin{eqnarray}
\label{eq:value_func}
&&\min_{G_i}\max_{D_i}\max_{\hat{D}_{i}}V(G_i,D_i,\hat{D}_{i})\\
&&=GAN\_loss_i + \lambda Council\_loss_i. \nonumber
\end{eqnarray}
This equation is a weighted sum of the adversarial loss  $GAN\_Loss_i$ (of $D_i$), as defined in~\cite{Mao2017LeastNetworks}, and the $Council\_loss_i$ (of $\hat{D_i}$) from Equation~\eqref{eq:loss_council}.
 $\lambda$ controls the importance of looking more "real" or more inline with the other generators.
High values will result in more similar images, whereas low values will require less agreement and result in higher diversity between the generated images.

\noindent
{\bf Focus map.}
For some applications, it is preferable to focus on specific areas of the image and modify only them, leaving the rest of the image untouched.
This can be easily accommodated into our general scheme, without changing the architecture.

The idea is to let the generator produce not only an image, but also an associated focus map, which essentially segments the learned objects in the domain from the background.
All that is needed is to add a fourth channel, $mask_i$, to the generator, which would generate values in the range~$[ 0,1]$.
These values can be interpreted as the likelihood of a pixel to belong to the background (or to an object).
To realize this, Equation~\eqref{eq:value_func} becomes
\begin{eqnarray}
\label{eq:value_func_focus}
&&\min_{G_i}\max_{D_i}\max_{\hat{D}_{i}}V(G_i,D_i,\hat{D}_{i})\\
&&=GAN\_loss_i + \lambda_1 Council\_loss_i + \lambda_2 Focus\_loss_i, \nonumber
\end{eqnarray}
where
\begin{eqnarray}
 \label{eq:loss_focus}
   Focus\_loss_i &=& \delta \big(\sum_{k}{mask_i[k]} \big)^2 \\
   &+& \sum_{k} \frac{1}{|mask_i[k]-0.5|+\epsilon}.  \nonumber
\end{eqnarray}

\begin{figure}
  \centering
    \begin{tabular}{ccccc}
    \includegraphics[height=0.085\textwidth]{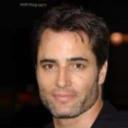}&
    \includegraphics[height=0.085\textwidth]{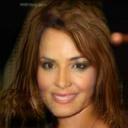}&
    \includegraphics[height=0.085\textwidth]{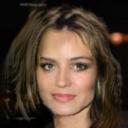}&   \includegraphics[height=0.085\textwidth]{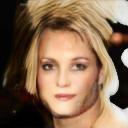}&
    \includegraphics[height=0.085\textwidth]{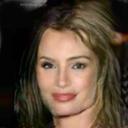}\\
    \includegraphics[height=0.085\textwidth]{Images/abliton_comp/input_ablation_exp_only_gan}&
    \includegraphics[height=0.085\textwidth]{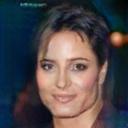}&
    \includegraphics[height=0.085\textwidth]{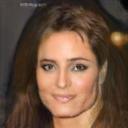}&   \includegraphics[height=0.085\textwidth]{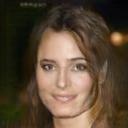}&
    \includegraphics[height=0.085\textwidth]{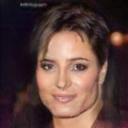}\\
    \includegraphics[height=0.085\textwidth]{Images/abliton_comp/input_ablation_exp_only_gan}&
    \includegraphics[height=0.085\textwidth]{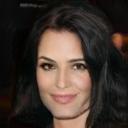}&
    \includegraphics[height=0.085\textwidth]{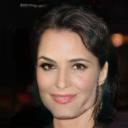}&   \includegraphics[height=0.085\textwidth]{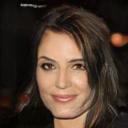}&
    \includegraphics[height=0.085\textwidth]{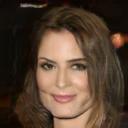}\\
    input& member1 & member2 & member3 & member4 
    \end{tabular}    
  \caption{{\bf Importance of the loss function components.} 
    This figure shows the results generated by the four council members for the male-to-female application, after $100K$ iterations.
    {\bf Top:} Using the $Focus\_loss$  (jointly with the classical $GAN\_loss$) generates nice images from the target domain, which are not necessarily related to the given image. 
    {\bf Middle:} Using the $Council\_loss$ instead, relates between the input and the output faces, but might change the environment (background). 
    {\bf Bottom:} Our loss, which combines the above losses, both relates the input and the output faces and focuses only on facial modifications. }
\label{fig:losses}
\end{figure}

In Equation~\eqref{eq:loss_focus}, $mask_i[k]$ is the value of the $4^{th}$ channel for pixel $k$.
The first term attempts to minimize the size of the focus mask, i.e. make it focus solely on the object.
The second term is in charge of segmenting the image into an object and a background ($1$ or $0$).
This is done in order to avoid generating semi-transparent pixels.
In our implementation $\epsilon=0.01$.
The result is normalized by the image size.
The values of $\lambda_1$ and $\lambda_2$ are application-dependent and will be defined for each application in Section~\ref{sec:implementation}.

Figure~\ref{fig:losses} illustrates the importance of the various losses.
If only the $Focus\_loss$ (jointly with the $GAN\_loss$) is used, the faces of the input and the output are completely unrelated, though the quality of the images is good and the background does not change in most cases.
Using only the $Council\_loss$, the faces of the input and the output are nicely related, but the background might change.
Our loss, which combines the above losses, produces the best results.

We note that this idea of adding a $4^{th}$ channel, which makes the generator focus on the proper areas of the image, can be used in other GAN architectures.
It is not limited to our proposed council architecture.

\section{Experiments}
\label{sec:experiments}

\subsection{Experiment setup}
\label{subsec:setup}
We applied our council GAN to several challenging image-to-image translation tasks (Section~\ref{subsec:results}).

\noindent
{\bf Baseline models.}
Depending on the application, we compare our results to those of some state-of-the-art models, including CycleGAN~\cite{Zhu2017UnpairedNetworks}, 
MUNIT~\cite{Huang2018MultimodalTranslation}, DRIT++~\cite{Lee_2018_ECCV,DRIT_plus}, U-GAT-IT~\cite{DBLP:journals/corr/abs-1907-10830}, StarGAN~\cite{StarGAN},  Fixed-PointGAN~\cite{siddiquee2019learning}.
These methods are unsupervised and use cycle constraints.
Out of these methods, MUNIT~\cite{Huang2018MultimodalTranslation} and DRIT++~\cite{Lee_2018_ECCV,DRIT_plus} are multi-modal and generate several results for a given image.
The others produce a single result.
Furthermore,  StarGAN~\cite{StarGAN} performs translation between multiple domains.

\noindent
{\bf Datasets.}
We evaluated the performance of our system on the following datasets.

{\bf CelebA~\cite{liu2015faceattributes}.}
This dataset contains $202,599$ face images of celebrities, each annotated with $40$ binary attributes. 
We focus on two attributes: (1) the gender attribute and (2) with/without glasses attribute.
The training dataset contains $68,261$ (/$10,521$) images of males (/with glasses)  and $94,509$ (/$152,249$)  images of females  (/without glasses).
The test dataset consists of $16,173$ (/$2,672$) males  (/with glasses) and $23,656$  (/$37,157$) females (/without glasses).

{\bf selfie2anime~\cite{DBLP:journals/corr/abs-1907-10830}.}
The size of the training dataset is $3,400$ selfie images and $3,400$ anime images.
The size of the test dataset is $100$ selfie images and $100$ anime images.

\noindent
{\bf Training.}
All models were trained using Adam~\cite{kingma2014adam} with $\beta_1 = 0.5$ and $\beta_1 = 0.999$. 
For data augmentation we flipped the images horizontally with a probability of $0.5$.

For the selfie/anime dataset , where the number of images is small, we augmented the data also with color jittering with up to $hue=0.15$, random Grayscale with a probability of $0.25$, random Rotation with up to $35^\circ$, random translation of up to $0.1$ of the image, and with random perspective with distortion scale of $0.35$  with a probability of $0.5$.
On the last $100K$ iterations we trained only on the original data, without augmentation.
We performed one generator update after a number of discriminator updates that is equal to the size of the council.
The batch size was set to $3$ for all experiments. 
We trained all models with a learning rate of $0.0001$, where the learning rate drops by a factor of $0.5$ after every $100,000$ iterations. 
The focus and council losses were added after $10,000$ iterations.

\noindent
{\bf Computational cost.}
The training takes about twice the time comparable to CycleGAN, when the council members
run sequentially on the same GPU.
The longer time is due to (1) having $4$ members (2) a longer iteration of the council-discriminator, and (3) twice as many iterations needed to reach an agreement. 
It only takes twice as long since we avoid learning of the reverse side (e.g. from anime to selfie). 
To accelerate the computation, the members could be run in parallel or a smaller council could be used.

\begin{figure*}
  \centering
    \begin{tabular}{ccccccccc}
   
    \includegraphics[height=0.09\textwidth]{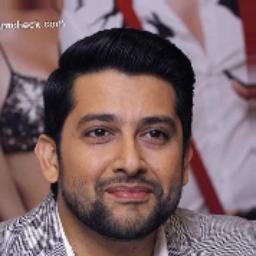}&
    \includegraphics[height=0.09\textwidth]{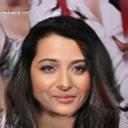}&
    \includegraphics[height=0.09\textwidth]{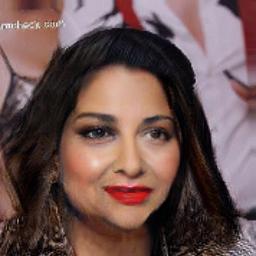}&
    \includegraphics[height=0.09\textwidth]{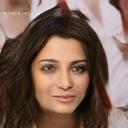}&
    \includegraphics[height=0.09\textwidth]{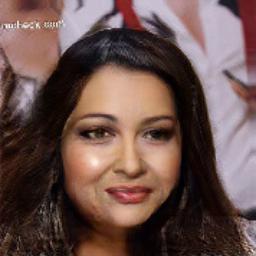}&
    \includegraphics[height=0.09\textwidth]{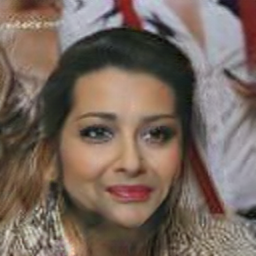}&
    \includegraphics[height=0.09\textwidth]{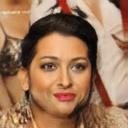}&
    \includegraphics[height=0.09\textwidth]{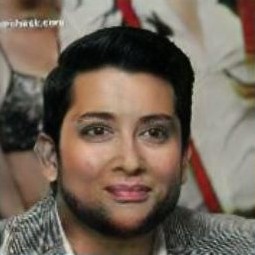}&
    \includegraphics[height=0.09\textwidth]{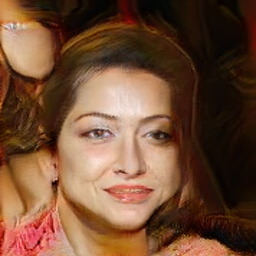}\\

    \includegraphics[height=0.09\textwidth]{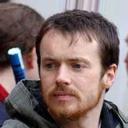}&
    \includegraphics[height=0.09\textwidth]{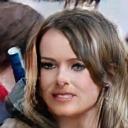}&
    \includegraphics[height=0.09\textwidth]{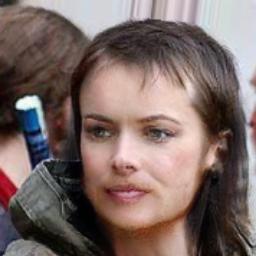}&
    \includegraphics[height=0.09\textwidth]{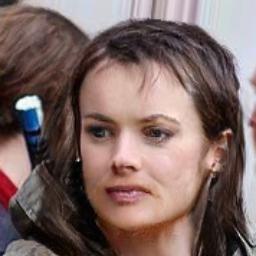}&
    \includegraphics[height=0.09\textwidth]{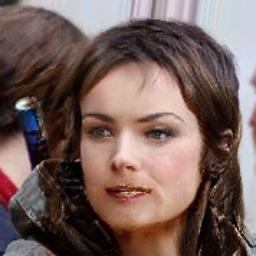}&
    \includegraphics[height=0.09\textwidth]{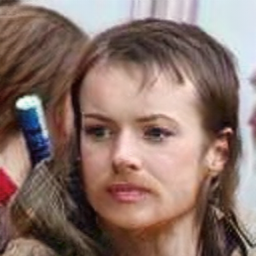}&
    \includegraphics[height=0.09\textwidth]{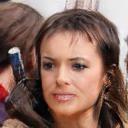}&
    \includegraphics[height=0.09\textwidth]{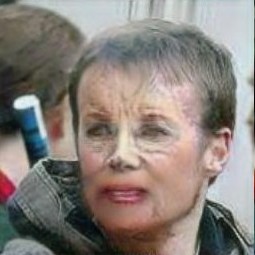}&
    \includegraphics[height=0.09\textwidth]{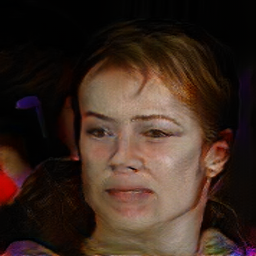}\\

    \includegraphics[height=0.09\textwidth]{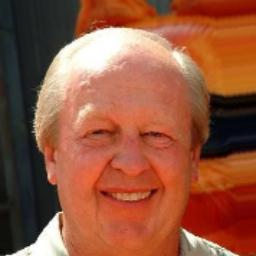}&
    \includegraphics[height=0.09\textwidth]{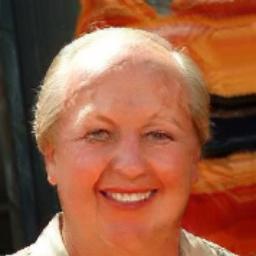}&
    \includegraphics[height=0.09\textwidth]{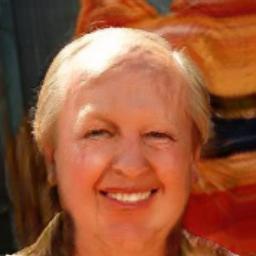}&
    \includegraphics[height=0.09\textwidth]{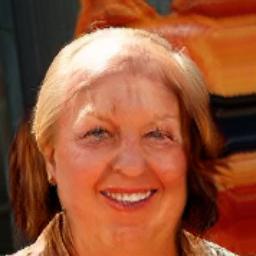}&
    \includegraphics[height=0.09\textwidth]{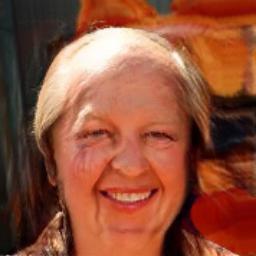}&
    \includegraphics[height=0.09\textwidth]{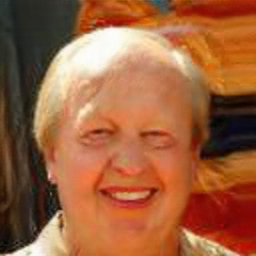}&
    \includegraphics[height=0.09\textwidth]{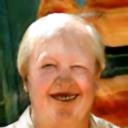}&
    \includegraphics[height=0.09\textwidth]{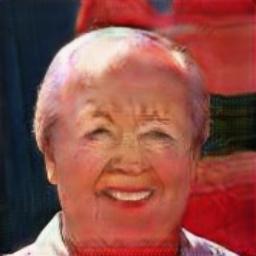}&
    \includegraphics[height=0.09\textwidth]{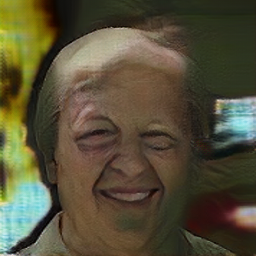}\\

    \includegraphics[height=0.09\textwidth]{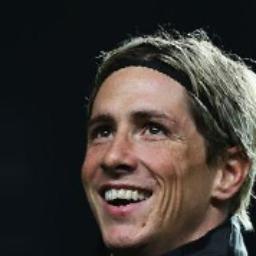}&
    \includegraphics[height=0.09\textwidth]{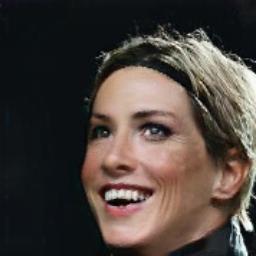}&
    \includegraphics[height=0.09\textwidth]{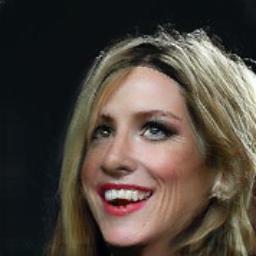}&
    \includegraphics[height=0.09\textwidth]{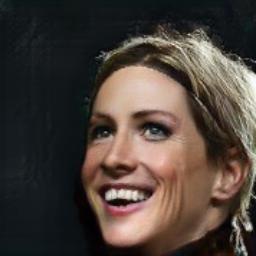}&
    \includegraphics[height=0.09\textwidth]{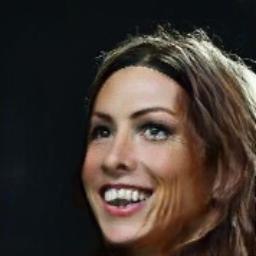}&
    \includegraphics[height=0.09\textwidth]{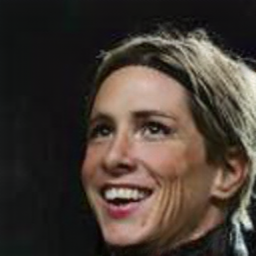}&
    \includegraphics[height=0.09\textwidth]{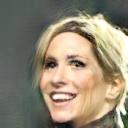}&
    \includegraphics[height=0.09\textwidth]{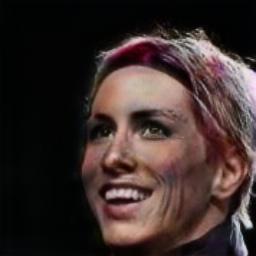}&
    \includegraphics[height=0.09\textwidth]{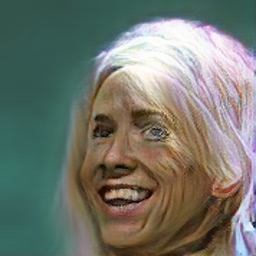}\\

    \includegraphics[height=0.09\textwidth]{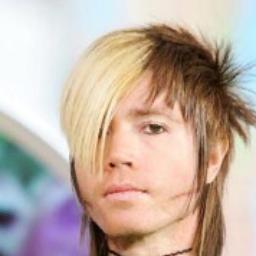}&
    \includegraphics[height=0.09\textwidth]{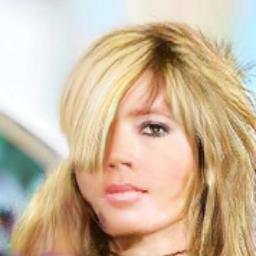}&
    \includegraphics[height=0.09\textwidth]{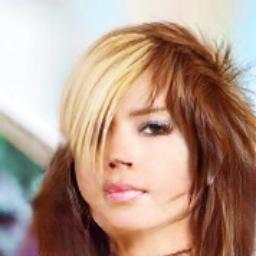}&
    \includegraphics[height=0.09\textwidth]{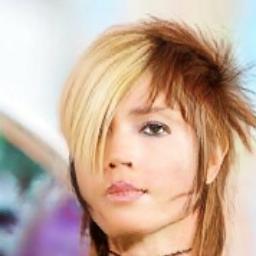}&
    \includegraphics[height=0.09\textwidth]{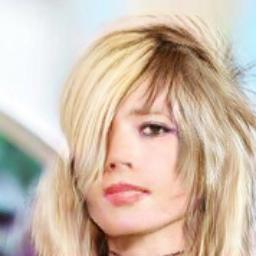}&
    \includegraphics[height=0.09\textwidth]{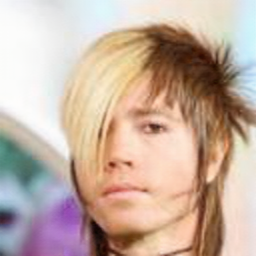}&
    \includegraphics[height=0.09\textwidth]{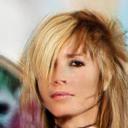}&
    \includegraphics[height=0.09\textwidth]{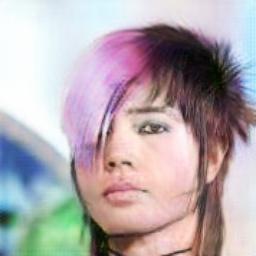}&
    \includegraphics[height=0.09\textwidth]{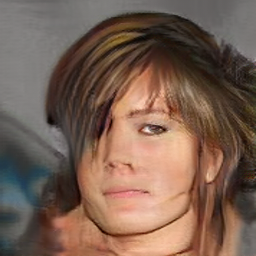}\\

    \includegraphics[height=0.09\textwidth]{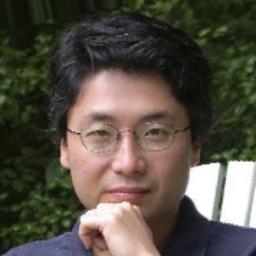}&
    \includegraphics[height=0.09\textwidth]{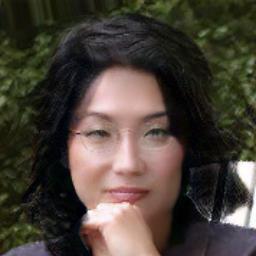}&
    \includegraphics[height=0.09\textwidth]{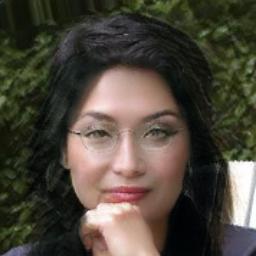}&
    \includegraphics[height=0.09\textwidth]{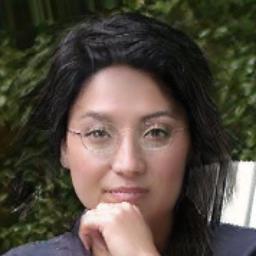}&
    \includegraphics[height=0.09\textwidth]{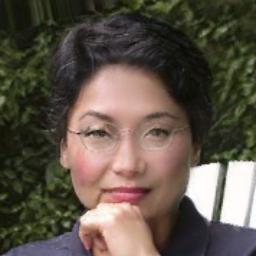}&
    \includegraphics[height=0.09\textwidth]{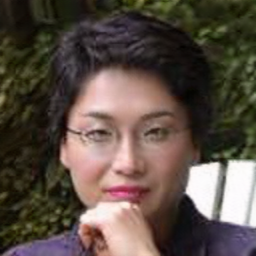}&
    \includegraphics[height=0.09\textwidth]{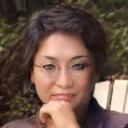}&
    \includegraphics[height=0.09\textwidth]{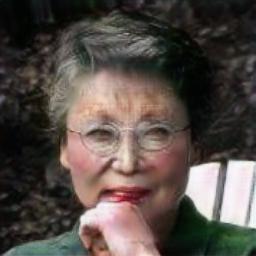}&
    \includegraphics[height=0.09\textwidth]{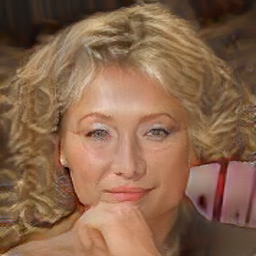}\\
     input& ours-1 & ours-2 & ours-3 & ours-4 & cycleGAN & MUNIT & StarGAN & DRIT++\\
     &  &  &  & & \cite{Zhu2017UnpairedNetworks} & \cite{Huang2018MultimodalTranslation} & \cite{StarGAN} & \cite{DRIT_plus,Lee_2018_ECCV}
    \end{tabular}
  \caption{{\bf Male-to-female translation.} 
    Our results  are more "feminine" than those generated by other state-of-the-art methods, while still preserving the main facial features of the input images.
    }
\label{fig:male-female}
\end{figure*}

\noindent
{\bf Evaluation.}
We verify our results both qualitatively and quantitatively.
For the latter, we use two common measures:
(1) the {\em Frechet Inception Distance score (FID)}~\cite{heusel2017gans}, which calculates the distance between the feature vectors of the real and the generated images;
(2) the {\em Kernel Inception Distance (KID)}~\cite{binkowski2018demystifying}, which improves on FID and measures GAN convergence.

\subsection{Experimental results}
\label{subsec:results}

\begin{figure*}
  \centering
    \begin{tabular}{ccccccccc}

    \includegraphics[height=0.09\textwidth]{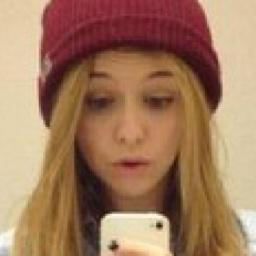}&
    \includegraphics[height=0.09\textwidth]{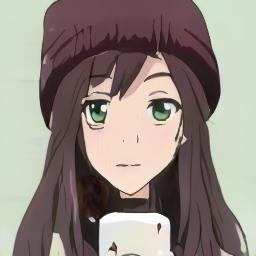}&
    \includegraphics[height=0.09\textwidth]{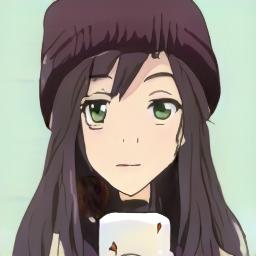}&
    \includegraphics[height=0.09\textwidth]{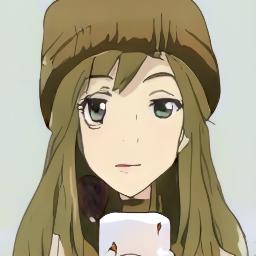}&
    \includegraphics[height=0.09\textwidth]{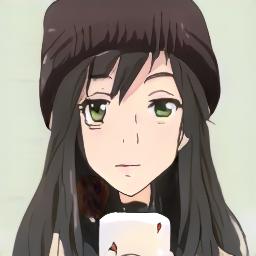}&
    \includegraphics[height=0.09\textwidth]{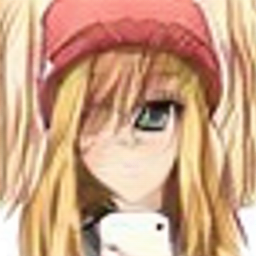}&
    \includegraphics[height=0.09\textwidth]{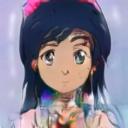}&
    \includegraphics[height=0.09\textwidth]{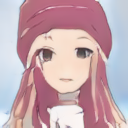}&
    \includegraphics[height=0.09\textwidth]{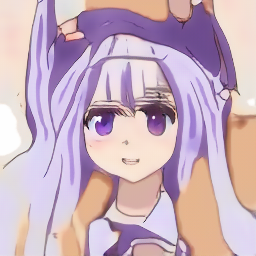}\\

    \includegraphics[height=0.09\textwidth]{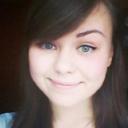}&
    \includegraphics[height=0.09\textwidth]{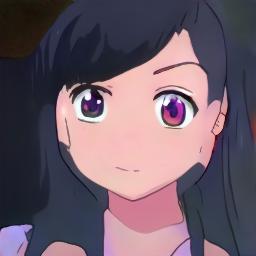}&
    \includegraphics[height=0.09\textwidth]{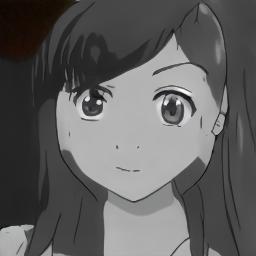}&
    \includegraphics[height=0.09\textwidth]{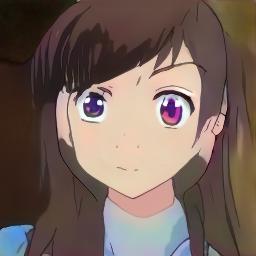}&
    \includegraphics[height=0.09\textwidth]{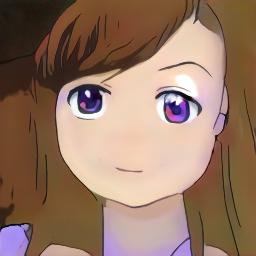}&
    \includegraphics[height=0.09\textwidth]{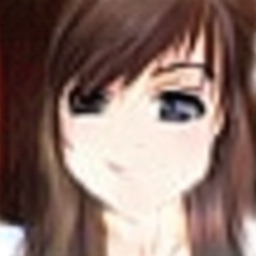}&
    \includegraphics[height=0.09\textwidth]{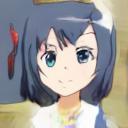}&
    \includegraphics[height=0.09\textwidth]{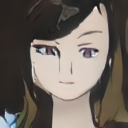}&
    \includegraphics[height=0.09\textwidth]{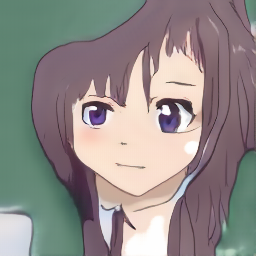}\\
    
    \includegraphics[height=0.09\textwidth]{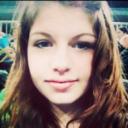}&
    \includegraphics[height=0.09\textwidth]{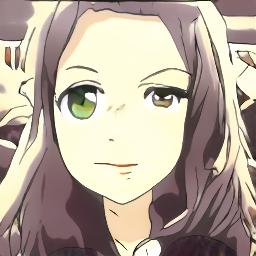}&
    \includegraphics[height=0.09\textwidth]{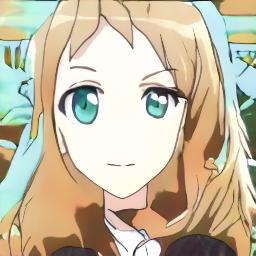}&
    \includegraphics[height=0.09\textwidth]{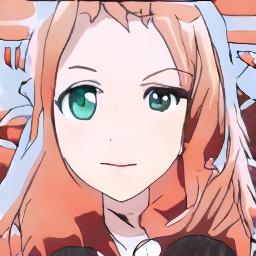}&
    \includegraphics[height=0.09\textwidth]{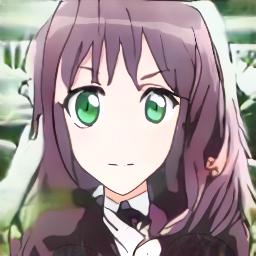}&
    \includegraphics[height=0.09\textwidth]{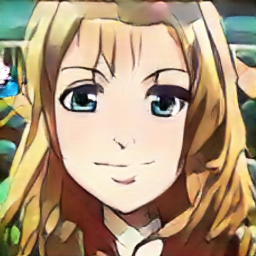}&
    \includegraphics[height=0.09\textwidth]{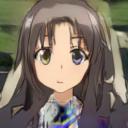}&
    \includegraphics[height=0.09\textwidth]{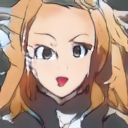}&
    \includegraphics[height=0.09\textwidth]{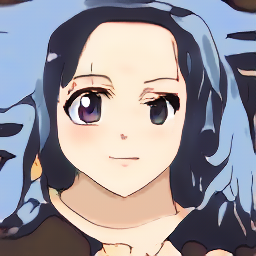}\\

    \includegraphics[height=0.09\textwidth]{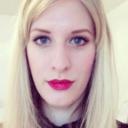}&
    \includegraphics[height=0.09\textwidth]{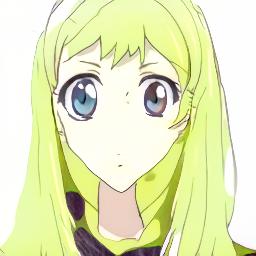}&
    \includegraphics[height=0.09\textwidth]{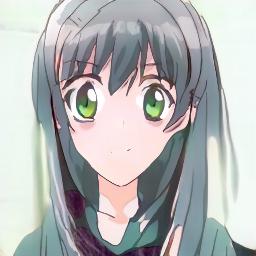}&
    \includegraphics[height=0.09\textwidth]{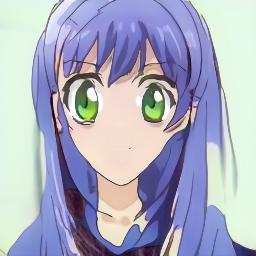}&
    \includegraphics[height=0.09\textwidth]{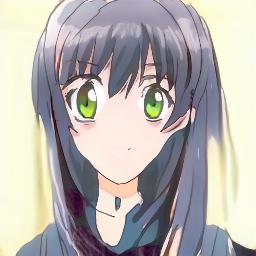}&
    \includegraphics[height=0.09\textwidth]{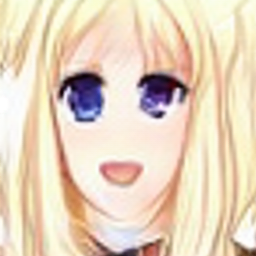}&
    \includegraphics[height=0.09\textwidth]{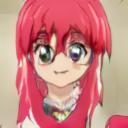}&
    \includegraphics[height=0.09\textwidth]{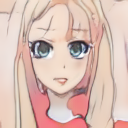}&
    \includegraphics[height=0.09\textwidth]{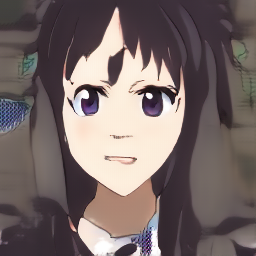}\\

    \includegraphics[height=0.09\textwidth]{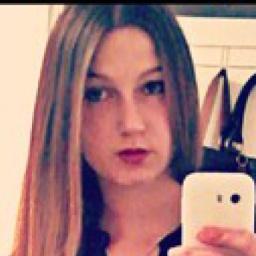}&
    \includegraphics[height=0.09\textwidth]{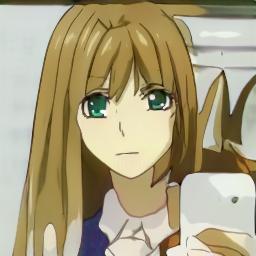}&
    \includegraphics[height=0.09\textwidth]{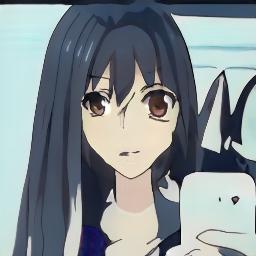}&
    \includegraphics[height=0.09\textwidth]{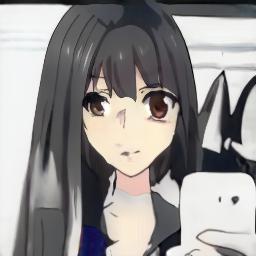}&
    \includegraphics[height=0.09\textwidth]{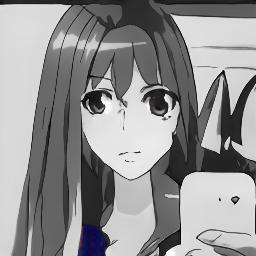}&
    \includegraphics[height=0.09\textwidth]{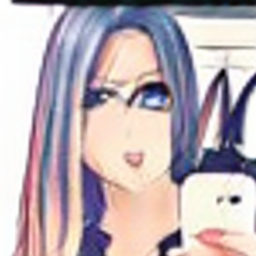}&
    \includegraphics[height=0.09\textwidth]{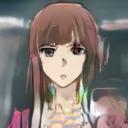}&
    \includegraphics[height=0.09\textwidth]{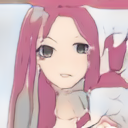}&
    \includegraphics[height=0.09\textwidth]{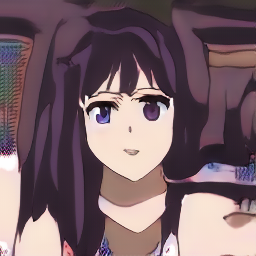}\\

    \includegraphics[height=0.09\textwidth]{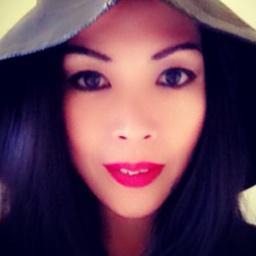}&
    \includegraphics[height=0.09\textwidth]{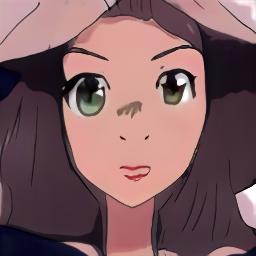}&
    \includegraphics[height=0.09\textwidth]{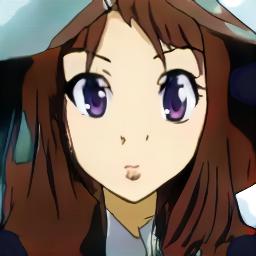}&
    \includegraphics[height=0.09\textwidth]{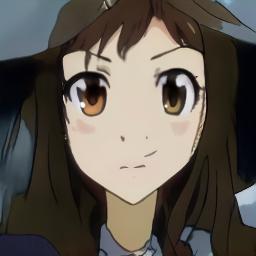}&
    \includegraphics[height=0.09\textwidth]{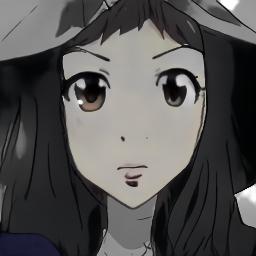}&
    \includegraphics[height=0.09\textwidth]{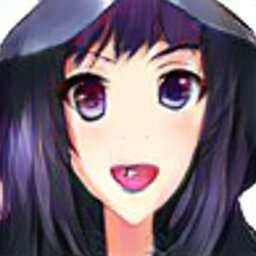}&
    \includegraphics[height=0.09\textwidth]{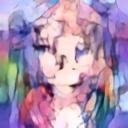}&
    \includegraphics[height=0.09\textwidth]{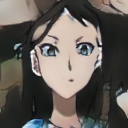}&
    \includegraphics[height=0.09\textwidth]{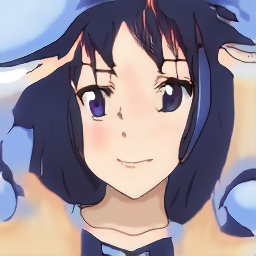}\\

    input& ours-1 & ours-2 & ours-3 & ours-4 & cycleGAN & MUNIT &  U-GAT-IT & DRIT++\\
    &  &  &  &  & \cite{Zhu2017UnpairedNetworks} & \cite{Huang2018MultimodalTranslation} &  \cite{DBLP:journals/corr/abs-1907-10830} & \cite{DRIT_plus,Lee_2018_ECCV}
    \end{tabular}    
  \caption{{\bf Selfie-to-anime translation.} 
    Our results preserve the structure of the face in the input image, while generating the characteristic features of anime, such as the large eyes. }
\label{fig:selfie2anime}
\end{figure*}

\noindent
{\bf Experimental results for male-to-female translation.}
Given an image of a male face, the goal is to generate a female face, which resembles
the  male face~\cite{Almahairi2018AugmentedData, lu2018attribute}. 
As explained in~\cite{Almahairi2018AugmentedData}, three features make this translation task challenging:
(i) There is no predefined correspondence in real data of each domain. 
(ii) The relationship is many-to-many between domains, as many male-to-female mappings are possible.
(iii) Capturing realistic variations in generated faces requires transformations that go beyond simple color and texture changes.

Figure~\ref{fig:male-female} compares our results, generated by a council of four members, to those of~\cite{StarGAN,Huang2018MultimodalTranslation,DRIT_plus,Zhu2017UnpairedNetworks}. 
Note that each of the council member may generate multiple results, depending on the random entropy vector.
We observe that our generated females are more "feminine" (e.g., the beards completely disappear and the haircuts are longer), while still preserving the main features of the source male face and resemble it. 
This can be attributed to the fact that we do not use a cycle to go from a male to a female and back, and thus we do not need to preserve any 
masculine features.
More examples are given in the supplementary materials

Table~\ref{tbl:male2female} summarized our quantitative results, where our results are randomly chosen from those generated by the different members of the council.
Our results outperform those of other methods in both evaluation metrics.

\begin{table}[tb]
    \centering
    \begin{tabular}{|c|c|c|}
    \hline
    & FID &  KID\\ 
    \hline
    CycleGAN~\cite{Zhu2017UnpairedNetworks} &20.91  &  0.0012 \\
    \hline
    MUINT~\cite{Huang2018MultimodalTranslation}  &   19.88 & 0.0013 \\
    \hline
    starGAN~\cite{StarGAN} &   35.50 &  0.0027 \\ 
    \hline
    DIRT++~\cite{DRIT_plus,Lee_2018_ECCV}  &  26.24 &  0.0016 \\ 
    \hline
    Council & {\bf 18.85} & {\bf  0.0010} \\
    \hline

    \end{tabular}
    \caption{{\bf Quantitative results for male-to-female translation.}  
    Our council generates results that outperform other SOTA results.
    For both measures, the lower the better.}
    \label{tbl:male2female}
\end{table}

\noindent
{\bf Experimental results for selfie-to-anime translation.}
Given an image of a human face, the goal is to generate an appealing anime, which resembles
the  human.
This is a challenging task, as not only the style differs, but also the geometric structure of the input and the output  greatly varies (e.g. the size of the eyes).
This might lead to mismatching of the structures, which would lead to distortions and visual artifacts.
This difficulty is added to the three challenges mentioned in the previous application: the lack of predefined correspondence of the domains, the  many-to-many relationship, and going beyond color and texture.

Figure~\ref{fig:selfie2anime} shows our results using a council of four. 
Our generated anime images are quite often better resemble the input in terms of expression and face structure (i.e., the shape of the chin) than  those of~\cite{Huang2018MultimodalTranslation,DBLP:journals/corr/abs-1907-10830,Lee_2018_ECCV,DRIT_plus,Zhu2017UnpairedNetworks}.
This can be explained by the fact that it is easier for the council members to "agree" on features that exist in the input.
Table~\ref{tbl:selfie2anime} shows quantitative results.
It can be seen that our results outperform or are competitive with those of other methods in both evaluation metrics.

\begin{table}[tb]
    \centering
    \begin{tabular}{ |c|c|c|}
    \hline
    & FID  & KID\\ 
    \hline
    CycleGAN~\cite{Zhu2017UnpairedNetworks} & 149.38   & 0.0056  \\
    \hline
    MUINT~\cite{Huang2018MultimodalTranslation} & 131.69  & 0.0057  \\
    \hline
    U-GAT-IT~\cite{DBLP:journals/corr/abs-1907-10830}  &  115.11  &  0.0043  \\ 
    \hline
    DIRT++~\cite{DRIT_plus,Lee_2018_ECCV}  & 109.22  & {\bf 0.0020}  \\ 
    \hline
    Council  & {\bf 101.39 } & {\bf 0.0020 }  \\
    \hline
    \end{tabular}
    \caption{{\bf Quantitative results for selfie-to-anime translation.} Our results outperform those of other methods when FID is considered and are competitive for KID. }
    \label{tbl:selfie2anime}
\end{table}

\noindent
{\bf Experimental results for glasses removal.}
Given an image of a person with glasses, the goal is to generate an image of the same person, but with the glasses removed. 
While in the previous application, the whole image changes, here the challenge is to modify only a certain part of the face and leave the rest of the image untouched.

\begin{figure}
  \centering
    \begin{tabular}{cccc}

    \includegraphics[height=0.09\textwidth]{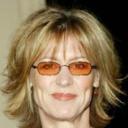}&
    \includegraphics[height=0.09\textwidth]{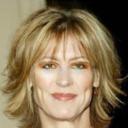}&
    \includegraphics[height=0.09\textwidth]{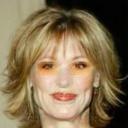}&
    \includegraphics[height=0.09\textwidth]{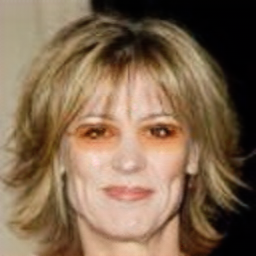}\\

    \includegraphics[height=0.09\textwidth]{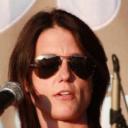}&
    \includegraphics[height=0.09\textwidth]{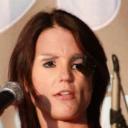}&
    \includegraphics[height=0.09\textwidth]{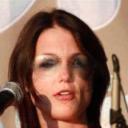}&
    \includegraphics[height=0.09\textwidth]{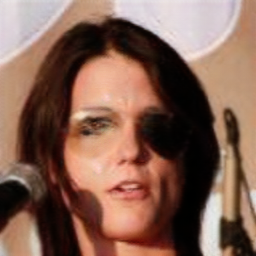}\\

    \includegraphics[height=0.09\textwidth]{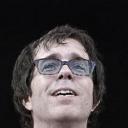}&
    \includegraphics[height=0.09\textwidth]{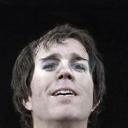}&
    \includegraphics[height=0.09\textwidth]{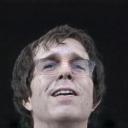}&
    \includegraphics[height=0.09\textwidth]{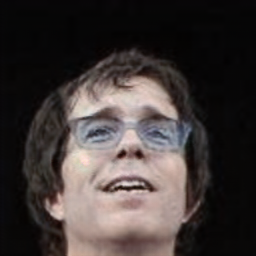}\\

    \includegraphics[height=0.09\textwidth]{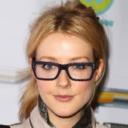}&
    \includegraphics[height=0.09\textwidth]{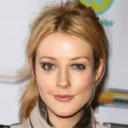}&
    \includegraphics[height=0.09\textwidth]{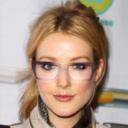}&
    \includegraphics[height=0.09\textwidth]{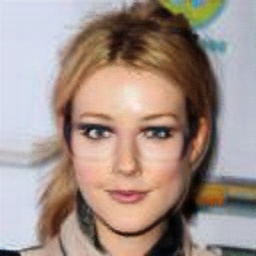}\\

    \includegraphics[height=0.09\textwidth]{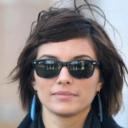}&
    \includegraphics[height=0.09\textwidth]{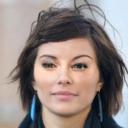}&
    \includegraphics[height=0.09\textwidth]{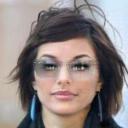}&
    \includegraphics[height=0.09\textwidth]{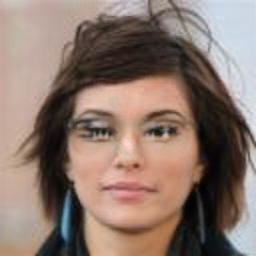}\\
    
    \includegraphics[height=0.09\textwidth]{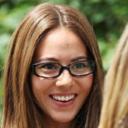}& 
    \includegraphics[height=0.09\textwidth]{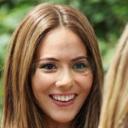}&
    \includegraphics[height=0.09\textwidth]{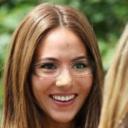}&
    \includegraphics[height=0.09\textwidth]{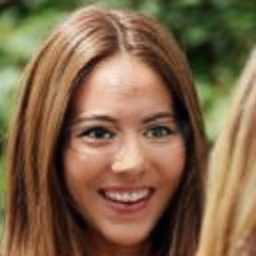}\\

    input& ours & Fixed-Point & cycleGAN\\
    &  & ~\cite{siddiquee2019learning} & ~\cite{Zhu2017UnpairedNetworks}
    \end{tabular}    
  \caption{{\bf Glasses removal.} We show a single result per input, since multi-modality is irrelevant for this application.
  Our generated images remove the glasses almost completely, whereas traces are left in~\cite{siddiquee2019learning}'s and in ~\cite{Zhu2017UnpairedNetworks}'s results. }
\label{fig:glasses}
\end{figure}

Figure~\ref{fig:glasses} compares our results (using a council of four) to those of~\cite{siddiquee2019learning}, which shows results for this application, as well as to~\cite{Zhu2017UnpairedNetworks}. 
Our generated images leave considerably less traces of the removed glasses.
Again, this can be attributed lack of the cycle constraint.
Table~\ref{tbl:glasses} provides quantitative results.
For this application as well, our council manages to outperform other methods and address the challenge of removing large objects.

\begin{table}[tb]
    \centering
    \begin{tabular}{ |c|c|c|}
    \hline
    \ & FID  & KID \\ 
    \hline
    cycleGAN~\cite{Zhu2017UnpairedNetworks} & 50.72   & 0.0038  \\
    \hline
    Fixed-point GAN~\cite{siddiquee2019learning} & 55.26  & 0.0041 \\
    \hline
    Council  & {\bf 36.38  } & {\bf 0.0026 }  \\
    \hline
    \end{tabular}
    \caption{{\bf Quantitative results of glasses removal.}  
    The results of our council outperform state-of-the-art results}
    \label{tbl:glasses}
\end{table}

\section{Implementation}
\label{sec:implementation}
Our code is based on PyTorch; it is available at https://github.com/Onr/Council-GAN. 
We set the major parameters as follows:
$\alpha$, which controls diversity (Equation~\eqref{eq:loss_council}), is set to $0.8$.
$\delta$, which controls the size of the mask  (Equation~\eqref{eq:loss_focus}), is set to $0.001$.
$\lambda_1$ and $\lambda_2$ from Equation~\eqref{eq:value_func_focus} are set according to the applications:
in male to female $\lambda_1=0.2$ \& $\lambda_2=0.025$;
in selfie to anime $\lambda_1=0.5$ \& $\lambda_2=0$;
in glasses removal $\lambda_1=0.2$ \& $\lambda_2=0.2$.

\begin{figure}[htb]
  \centering
     \includegraphics[width=0.4\textwidth]{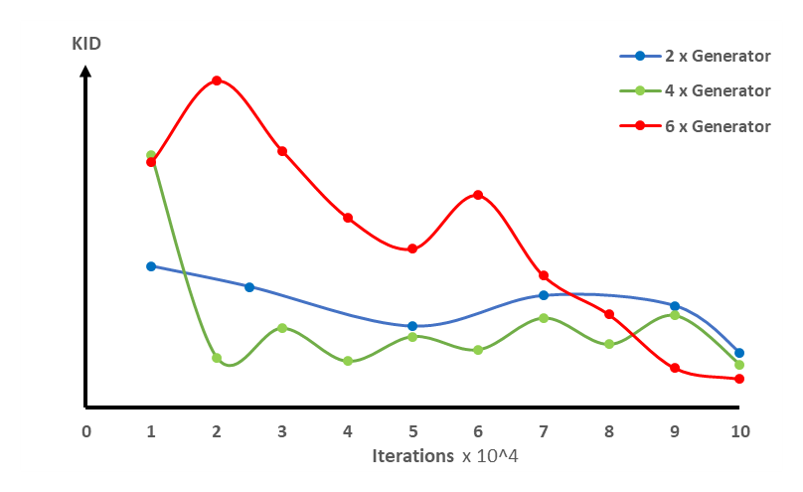}
    \caption{{\bf KID as a function of \# iterations.}  
    The more iterations, the better KID.
    Moreover, with more council members, model converges more slowly, yet the results improve.}
    \label{fig:members}
\end{figure}

Figure~\ref{fig:members} studies the influence of the number of members and the number of iterations on the quality of the results.
We focus on the male-to-female application, which is representative.
The fewer the number of members in the council, the faster the convergence is.
However, this comes at a price: the accuracy is worse.
Furthermore, it can be seen that the KID improves with iterations, as expected.

\begin{figure}[tb]
  \centering
    \begin{tabular}{cccc}
     \includegraphics[height=0.09\textwidth]{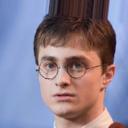}&
     \includegraphics[height=0.09\textwidth]{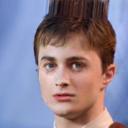} &
      \includegraphics[height=0.09\textwidth]{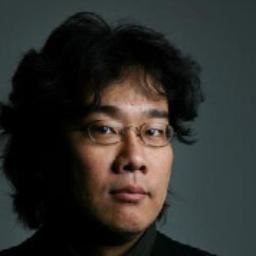}&
     \includegraphics[height=0.09\textwidth]{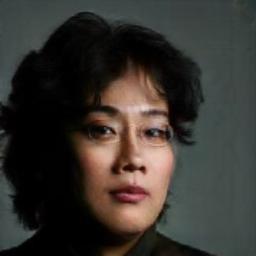}\\
     input & result & input & result \\
         \multicolumn{2}{c}{(a) glasses removal}&
         \multicolumn{2}{c}{(a) male to female}
   \end{tabular}
    \caption{{\bf Limitation.}  
    (a) When removing the glasses, the face also becomes more feminine. 
    (b) Conversely, when transforming a male to a female, the glasses may also be removed.
    This is attributed to high imbalance of the relevant features in the dataset.
    }
    \label{fig:limitation}
\end{figure}

\noindent
{\bf Limitations.}
Figure~\ref{fig:limitation} demonstrates a limitation of our method.
When removing the glasses, the face might also become more feminine.
This is attributed to the imbalance inherent to the dataset.
Specifically, the ratio of the men to women with glasses is $0.8$, whereas the ratio of men to women without glasses is only $0.4$.
The result of this imbalance in the target domain is that removing glasses also means becoming more feminine.
This problem can be solved by providing a dataset with an equal number of males and females with and without glasses.
Handling feature imbalance without changing the number of images in the dataset, is an interesting direction for future research.

\section{Conclusion}
\label{sec:conclusion}
This paper introduces the concept of a council of GANs---a novel approach to perform image-to-image translation between unpaired domains.
They key idea is to replace the widely-used cycle-consistency constraint by leveraging collaboration between GANs.
Council members assist each other to improve, each its own result.

Furthermore, the paper proposes an implementation of this concept and demonstrates its benefits for three challenging applications.
The members of the council generate several optional results for a given input.
They manage to remove large objects from the images, not to leave redundant traces from the input and to handle large shape modifications.
The results outperform those of SOTA algorithms both quantitatively and qualitatively.

\noindent
{\bf Acknowledgements:} We gratefully acknowledge the support of the Israel Science Foundation (ISF) 1083/18,  PMRI-Peter Munk Research Institute--Technion, and NVIDIA Corporation with the donation of the GPU.

\newpage
\bibliographystyle{ieee}
\small
\bibliography{citation.bib}

\end{document}